\documentclass[runningheads]{llncs}

\usepackage{eccv}

\usepackage{eccvabbrv}

\usepackage{graphicx}
\usepackage{booktabs}

\usepackage[accsupp]{axessibility}  % Improves PDF readability for those with disabilities.

\usepackage{hyperref}

\usepackage{orcidlink}

\usepackage{multirow} 
\usepackage{pifont}

\begin{document}

\title{Unified Embedding Alignment for Open-Vocabulary Video Instance Segmentation} 

\titlerunning{Unified Embedding Alignment for Open-Vocabulary VIS}

\author{Hao Fang\inst{1}\orcidlink{0000-0002-8846-8294} \and
Peng Wu\inst{1}\orcidlink{0000-0003-1137-0720} \and
Yawei Li\inst{2}\orcidlink{0000-0002-8948-7892} \and
Xinxin Zhang\inst{1}\orcidlink{0000-0001-6069-5391} \and
Xiankai Lu\inst{1}\thanks{Corresponding author: carrierlxk@gmail.com}\orcidlink{0000-0002-9543-6960}}

\authorrunning{H.~Fang et al.}

\institute{School of Software, Shandong University \and
Computer Vision Lab, ETH Zurich}

\maketitle

\begin{abstract}
Open-Vocabulary Video Instance Segmentation~(VIS) is attracting increasing attention due to its ability to segment and track arbitrary objects. However, the recent Open-Vocabulary VIS attempts obtained unsatisfactory results, especially in terms of generalization ability of novel categories. We discover that the domain gap between the VLM features (\eg,
CLIP) and the instance queries and the underutilization of temporal consistency are two central causes. To mitigate these issues, we design and train a novel Open-Vocabulary VIS baseline called OVFormer. OVFormer utilizes a lightweight module for unified embedding alignment between query embeddings and CLIP image embeddings to remedy the domain gap. Unlike previous \textit{image}-based training methods, we conduct \textit{video}-based model training and deploy a semi-online inference scheme to fully mine the temporal consistency in the video. Without bells and whistles, OVFormer achieves 21.9 mAP with a ResNet-50 backbone on LV-VIS, exceeding the previous state-of-the-art performance by 7.7. Extensive experiments on some Close-Vocabulary VIS datasets also demonstrate the strong zero-shot generalization ability of OVFormer (\texttt{+} 7.6 mAP on YouTube-VIS 2019, \texttt{+} 3.9 mAP on OVIS). Code is available at \href{https://github.com/fanghaook/OVFormer}{https://github.com/fanghaook/OVFormer}. 
  \keywords{Open-Vocabulary \and Video instance segmentation \and Embedding alignment \and Semi-Online}
\end{abstract}

\section{Introduction}
\label{sec:intro}
Open-Vocabulary Video Instance Segmentation (VIS)~\cite{wang2023towards,guo2023openvis} is an emerging vision task that aims to simultaneously classify, track, and segment arbitrary objects from an open set of categories in videos. Compared to VIS~\cite{yang2019video}, Open-Vocabulary VIS is much more challenging since it requires handling novel categories that have never appeared in the training set. Current state-of-the-art Open-Vocabulary VIS methods~\cite{wang2023towards,guo2023openvis} first use a query-based image instance segmentation backbone~\cite{cheng2022masked} to generate class-agnostic mask proposals for each frame, then perform open vocabulary classification with the frozen pre-trained vision-language models (VLMs)~\cite{radford2021learning}, and finally track object instances frame by frame using online paradigm~\cite{huang2022minvis}.
\begin{figure}
\begin{center}
\includegraphics[width=0.7\linewidth]{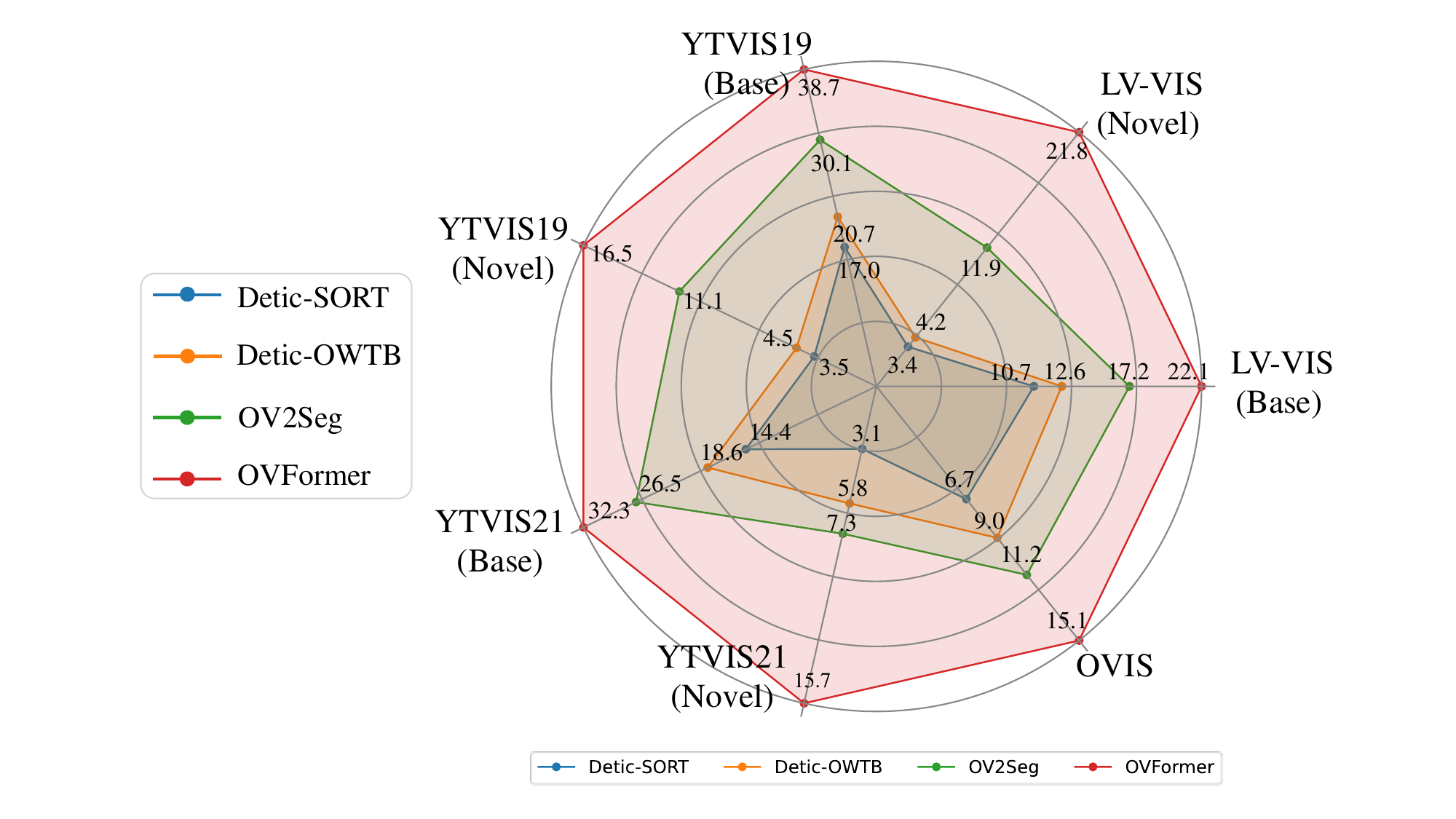}
\end{center}
\caption{The proposed OVFormer outperforms its counterparts significantly on the VIS datasets.}
\label{intro}
\end{figure}

While these existing Open-Vocabulary VIS solutions have shown significant promise, the following questions naturally arise: 1) achieving alignment between text and video in CLIP-based models, and 2) considering temporal consistency. 
\textbf{Firstly}, the open vocabulary method based on CLIP~\cite{radford2021learning} directly obtains the classification score of each instance by calculating the similarity between query embedding and text embedding. Due to the inconsistent training sample sources between VLM and VIS models, there is a huge domain gap between these two models, which is not conducive to text and video alignment. How to alleviate the domain gap has become the key to solving this problem. 
\textbf{Secondly}, the model training and inference in current methods are conducted on the image level rather than the video level,  which did not fully utilize rich spatio-temporal context in video. Thus, the classification and segmentation results are not temporal consistent for the same instance across frames. %This inspires us to design new training and inference paradigms to improve temporal consistency.

We have striking findings that the previous methods did not delve into these two issues in detail. Tackling these two issues can provide insights into open-vocabulary video instance segmentation model design, and motivate us to propose a novel elegant framework from \textit{unified embedding alignment view}, OVFormer. OVFormer seamlessly incorporates large pre-trained VLM and VIS model within an efficient video-level training regimen. Specifically, given the video clip, we first generate video-level instance queries to represent the entire clip. 
Second, we employ the cross-attention mechanism to reconcile the cross-model correlation between the VIS and VLM for unified embedding alignment. This simple and lightweight operation remedies the domain gap between query embeddings and CLIP text embeddings to facilitate open-vocabulary classification. 
Finally, to further exploit the temporal information in a video clip, we conduct video-based training and develop a semi-online inference regime to make our model flexible to various VIS settings. This also prompts us to perform embedded alignment at the video level instead of aligning each frame individually.

The proposed OVFormer brings significant performance improvements. OVFormer achieves 21.9 mAP with a ResNet-50 backbone on LV-VIS benchmark~\cite{wang2023towards}, exceeding the previous state-of-the-art performance by 7.7. The remarkable performance improvement on some Close-Vocabulary VIS datasets also demonstrates the strong zero-shot generalization ability of OVFormer (\texttt{+} 7.6 mAP on YouTube-VIS 2019, \texttt{+} 6.2 mAP on YouTube-VIS 2021, \texttt{+} 3.9 mAP on OVIS). The detailed performance comparison on base categories and novel categories is shown in \cref{intro}.
The contributions can be summarized as follows:
\begin{itemize}
\item We reveal that alignment between text and video and temporal consistency are two issues that puzzle the performance of Open-Vocabulary VIS. 
\item We propose a novel end-to-end open-vocabulary video instance segmentation framework, OVFormer, which represents a remarkable advancement in open-vocabulary capability for VIS through an elegant structure.
\item Our method performs unified embedding alignment between the instance query generation of VIS and the pre-trained VLM to facilitate obtaining feature-aligned class embeddings, thus enhancing the generalization ability of novel categories.
\item We conduct video-level training without modifying the architecture and deploy a flexible semi-online inference scheme, which improves the temporal consistency of video instance segmentation. 
\end{itemize}

\section{Related Work}
\label{sec:Related Work}
\noindent \textbf{Open-Vocabulary Video Instance Segmentation} is an emerging vision task that aims to simultaneously classify, track, and segment arbitrary objects from an open set of categories in videos. There are two popular works: Guo \etal~\cite{guo2023openvis} propose a two-stage Open-Vocabulary VIS framework, OVIS. They first adopt a query-based image instance segmentation model as the mask proposal network. In the second stage, they employ a pre-trained VLM to compute similarities between the obtained proposals and the text descriptions of all possible classes, then predict the most likely category. 
As a concurrent work, Wang \etal~\cite{wang2023towards} collect a Large-Vocabulary Video Instance Segmentation dataset (LV-VIS) for evaluating the generalization ability of Open-Vocabulary VIS methods. They also propose the first end-to-end Open-Vocabulary VIS benchmark, OV2Seg, which simplifies the intricate propose-reduce-association paradigm and attains long-term awareness with a Memory-Induced Transformer. Although the above methods have successfully introduced open-vocabulary into VIS, their generalization ability to recognize novel categories is unsatisfying.

\noindent \textbf{Video Instance Segmentation} aims to classify, segment, and track object instances from pre-defined training categories, which is more challenging than video object segmentation~\cite{yang2024scalable,lu2020deep,lu2020zero,lu2021segmenting}. Broadly, current VIS methods can be classified into \textit{online} and \textit{offline} methods. Online methods~\cite{yang2019video,lin2021video,huang2022minvis,wu2022defense,ying2023ctvis,li2023tcovis} tackle video instance segmentation on each frame, and apply post-processing steps to associate instances across frames explicitly.  For instance, MinVIS~\cite{huang2022minvis} and IDOL~\cite{wu2022defense} make use of the discriminative instance queries for matching between frames. 
In contrast, offline methods~\cite{wang2021end,hwang2021video,cheng2021mask2former,yang2022temporally,wu2022seqformer,heo2022vita,zhang2023dvis,fang2024learning} take a video clip as input and generate a sequence of instances of the video clip end-to-end. For example, Seqformer~\cite{wu2022seqformer} and VITA~\cite{heo2022vita} localize instances in each frame and learn a powerful representation of video-level instance queries, which is used to predict mask sequences on each frame. Offline methods leverage comprehensive temporal information from the whole clip and have the advantage of producing temporally consistent results. Nevertheless, as the number of video frames and instances significantly increases during the inference stage, the performance of offline methods experiences a notable decline. 

\noindent \textbf{Visual Recognition from Vision-Language Models}.  The vision-language models such as CLIP~\cite{radford2021learning} and ALIGN~\cite{jia2021scaling} pretrained on large-scale image-text pairs have shown transferability to various visual recognition tasks. Example of such tasks include object detection~\cite{gu2021open,ma2022open,zareian2021open,du2022learning}, image segmentation~\cite{ding2022decoupling,chen2023open,wu2023betrayed,zhou2022extract}, and video understanding~\cite{wang2021actionclip,li2023ovtrack}. Specifically, using the text embeddings from VLMs as the classifier for images can provide downstream models with open-vocabulary recognition capability. Typical text-image alignment approaches include knowledge distillation~\cite{gu2021open,ma2022open,li2023ovtrack}, regional text pre-training~\cite{zareian2021open,wang2021actionclip}, and prompt modeling~\cite{du2022learning}. Some open-vocabulary image segmentation methods~\cite{ding2022decoupling,chen2023open} further exploit image caption data~\cite{wu2023betrayed} and pseudo labels~\cite{zhou2022extract} to improve the performance. %For video understanding tasks, most works focus on designing VLM models’ temporal fusion or association in various settings, including action recognition~\cite{wang2021actionclip} and tracking~\cite{li2023ovtrack}.
Despite VLMs providing open-vocabulary capability for handling arbitrary
object classes in these works, the
problem of domain gap remains far from being 
resolved. Another common oversight in existing methods is the neglect of temporal structures among videos.  This work pursues a unified embedding alignment framework that addresses these fundamental limitations, providing a refreshing viewpoint on open-vocabulary video instance segmentation.
%directly using VLMs in downstream tasks causes several issues. As the differences among training data and annotations, there exists a domain gap between VLMs and downstream models, leading to poor generalization. Furthermore, unlike image-level visual recognition tasks, this paper focuses on addressing the domain gap between VLMs and VIS at the video level.

\section{Method}
\noindent\textbf{Motivation.}
We discover that the domain gap between the VLM features and the instance queries and the underutilization of temporal consistency are two central causes for the poor generalization performance of previous methods. These two observations motivate us to design the following two core steps: \textit{Unified Embedding Alignment} to remedy the domain gap of visual-textual features and \textit{Video-level Training Paradigm} to obtain the temporal consistency prediction. 

\begin{figure}[tb]
\begin{center}
\includegraphics[width=\linewidth]{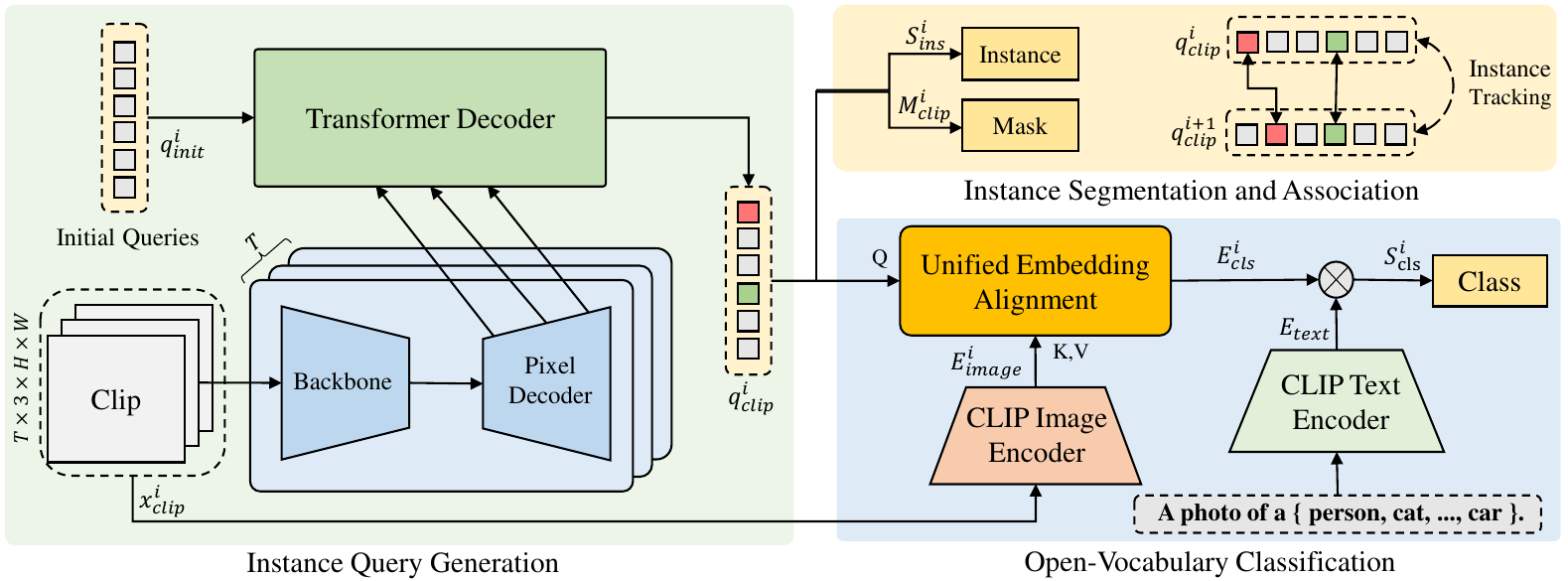}
\end{center}
\caption{Overview of OVFormer. Our proposed OVFormer consists of three modules: Instance Query Generation module generates video-level instance queries that represent the entire clip; Then, Open-Vocabulary Classification module performs unified embedding alignment between the instance queries and the image features of CLIP to generate feature-aligned class embeddings, then classify them with the text embeddings computed from a CLIP text encoder. Finally, Instance Segmentation and Association module generates and tracks the instances through bipartite matching.}
\label{fig: OVFormer}
\end{figure}

\noindent\textbf{Pipeline.}  As depicted in \cref{fig: OVFormer}, OVFormer consists of three modules: Initially, the input video clips are encoded as video-level instance queries by Instance Query Generation module with the query-based image instance segmentation model. Subsequently, the Open-Vocabulary Classification module takes both the original clips and video-level instance queries as input and performs unified embedding alignment to generate feature-aligned class embeddings. Next, this module classifies them with the text embeddings to realize open-vocabulary prediction. Simultaneously, Instance Segmentation and Association module generates and tracks the instances mask through bipartite matching.

\subsection{Instance Query Generation}
Given an input video clip $x_{clip}^i\in \mathbb{R}^{T\times 3\times H\times W}$ with 3 color channels and $T$ frames of resolution $H \times W$, we employ the Instance Query Generation module to produce instance queries per-clip. Specifically, the whole module consists of a backbone~\cite{he2016deep,liu2021swin} works for feature extraction and the pixel decoder gradually upsamples the features to generate multi-scale features and per-pixel embeddings:
\begin{equation}
\left(F^i_{enc},  E_{pixel}^i\right) = \Phi_{\textsc{enc}}\left(\operatorname{backbone}(x^i_{clip}),e_{pos}^{s}\right),
\label{equ:enc}
\end{equation}
where $i$ is the video clip index, $\Phi_{\textsc{enc}}$ denotes pixel decoder, $e_{pos}^{s}$ means spatial positional encoding~\cite{carion2020end}, $F^i_{enc}$ is multi-scale visual features, $E_{pixel}^i \in \mathbb{R}^{C \times T \times \frac{H}{S} \times \frac{W}{S}}$ denotes per-pixel embeddings, where $S$ is the stride of the feature map.

Instead of calculating instance queries for each frame independently as image-level training~\cite{wang2023towards} or generating mask proposals explicitly~\cite{guo2023openvis}, we use a set of video-level instance queries to represent the entire video clip, which further excavates spatio-temporal information and enhances temporal consistency of the instances. To obtain compatibility with image segmentation models, we further extend the positional encoding to the spatio-temporal domain by incorporating temporal positional encoding~\cite{cheng2021mask2former}:
\begin{equation}
e_{pos} = e_{pos}^t + e_{pos}^s \in \mathbb{R}^{C \times T \times H_l \times W_l},
\label{equ:pos}
\end{equation}
where $l$ is the Transformer decoder layer index, $e_{pos}$ is the final positional encoding, $e_{pos}^s\in \mathbb{R}^{C \times 1 \times H_l \times W_l}$ is the original spatial positional encoding, $e_{pos}^{t}\in \mathbb{R}^{C \times T \times 1 \times 1}$ is the corresponding temporal positional encoding. Both $e_{pos}^s$ and $e_{pos}^t$ are non-parametric sinusoidal positional encodings.

Then, the multi-scale features and initialized queries are taken into a Transformer decoder to obtain the video-level instance queries for the input video clip:
\begin{equation}
q_{clip}^{i}  = \Phi_{\textsc{dec}} (F^i_{enc}, q^i_{init}, e_{pos}) \in \mathbb{R}^{N\times C},
\label{equ:dec}
\end{equation}
where $\Phi_{\textsc{dec}}$ denotes Transformer decoder, $q_{init}^{i} \in \mathbb{R}^{N\times C}$ is random initialized instance queries. $q_{clip}^{i}$ is the video-level instance queries, $N$ denotes the instance number per clip.  We adopt a query-based image instance segmentation model~\cite{cheng2022masked} to implement the query generation network.

\subsection{Open-Vocabulary Classification}
After generating the instance queries for each video clip, we need to derive the open-vocabulary category and mask for each instance. 

\noindent\textbf{Unified Embedding Alignment.} 
To perform open-vocabulary classification in instance segmentation models,  current open-vocabulary VIS methods utilize pre-trained VLM, such as CLIP~\cite{radford2021learning} to replace the linear classification head of the original segmentation network and predict the novel category name via computing the similarity between the instance queries and VLM-encoded text features directly. 

However, the generated video instance queries $q_{clip}^{i}$ here are not aligned with the pre-trained VLM features due to the domain gap of the training sources, and directly calculating similarity with text features is not conducive to open-vocabulary classification. 
Therefore, we advocate unified embedding alignment between the instance queries and the image features of pre-trained VLM to generate feature-aligned class embeddings, greatly enhancing the generalization ability of novel categories. Our module is specially designed for video-level feature alignment, allowing for unified interaction between instance queries and CLIP features concatenated from multiple frames, which is fundamentally different from the image-level open vocabulary methods.

Specifically, we feed the video clip  $x_{clip}$ into the CLIP model and compute the plain visual feature through the CLIP image encoder:
\begin{equation}
  \label{equ:image}
  E_{image}^i  = \Phi_{image}(x^i_{clip}) \in \mathbb{R}^{T \times {C'}},
\end{equation}
where $\Phi_{image}$ denotes the CLIP image encoder, $E_{image}^i$ denotes CLIP image embeddings, $C'$ is the feature dimension output from the CLIP encoder.

Then, we employ a cross-attention mechanism to implement unified embedding alignment. Specifically,  we feed $E_{image}^i$ along with the video instance queries $q_{clip}^{i}$ (\cref{equ:dec}) into a cross-attention module~\cite{vaswani2017attention}. Here, instance queries serve as queries, while CLIP image embeddings take on the roles of both keys and values. The relationship is formulated as: 
\begin{align}
  \label{equ:cross-attention}
  Q_{clip}^{i} &= f_{\operatorname{mlp}} (q^i_{clip}) \in \mathbb{R}^{N \times {C'}}, \\
  V_i,K_i &= f_{\operatorname{proj}}(E_{image}^i) \in \mathbb{R}^{T \times {C'}}, \\
 E_{cls}^i &= \operatorname{softmax}(\frac{Q_{clip}^{i}K_i^\top}{\sqrt{C'}})V_i \in \mathbb{R}^{N \times {C'}},
 \end{align}
where $Q_{clip}^{i} $ is obtained by $q_{clip}^{i}$ through two MLP layers ($f_{\operatorname{mlp}}$) to maintain consistency with the dimension of CLIP feature dimension. $K_i$ and $V_i$ are derived from separate linear projections ($f_{\operatorname{proj}}$) of $E_{image}^i$, $E_{cls}^i$ denotes video-level class embeddings. This simple approach has consistently demonstrated its effectiveness in information fusion. Here, we leverage
this mechanism to integrate spatial-temporal information from the video instance queries with semantic content extracted from VLM model. In this way, the domain gap can be narrowed. 

\noindent\textbf{Open-Vocabulary Classification.} 
Once we obtain the enhanced video-level class embeddings, we perform open-vocabulary classification to assign the semantic label for each embedding. Specifically, we feed the vocabulary set into the CLIP text encoder to generate the text embedding as the category labels:
\begin{equation}
    \label{equ:text}
    E_{text} = \Phi_{text}(\text{``A photo of [category name]''}) \in \mathbb{R}^{K \times {C'}},
\end{equation}
where $\Phi_{text}$ is CLIP text encoder, $E_{text} $ denotes text embeddings, $K$ is the number of categories.

Finally, we multiply class embeddings and text embeddings to obtain the classification score for each video instance query:
\begin{equation}
  \label{equ:class}
 S_{cls}^i = \operatorname{softmax}(E_{cls}^i \cdot E_{text}^\top) \in \mathbb{R}^{N\times K},
\end{equation}
where $\operatorname{softmax}$ denotes a softmax operation, $\cdot$ denotes dot product operation, $S_{cls}^i$ is the classification scores of all instance queries. In this way, we can obtain the category name for each instance.

\subsection{Instance Segmentation and Association}
\noindent\textbf{Instance Segmentation.} 
In line with prior research,  we add a binary classifier as an instance head to predict objectiveness score for each instance: 
\begin{equation}
  \label{equ:instance}
 S_{ins}^i = \operatorname{InsHead}(q_{clip}^{i})  \in \mathbb{R}^{N\times 1},
\end{equation}
where $\operatorname{InsHead}$ is implemented with three MLP layers, $S_{ins}^i$ is the instance scores of all queries.

To obtain the predicted masks, we first transform the instance query into mask embedding. Then, we obtain the mask of each query by applying simple dot products between the mask embeddings and per-pixel embeddings:
\begin{align}
  \label{equ:mask}
E_{mask}^{i} &= \operatorname{MaskHead}(q_{clip}^i) \in \mathbb{R}^{N\times C},\\
\mathcal{M}_{clip}^i &=\operatorname{sigmoid}(E_{mask}^{i} \cdot E_{pixel}^{i}) \in \mathbb{R}^{N\times T\times \frac{H}{S}\times \frac{W}{S}}, 
\end{align}
where $\operatorname{MaskHead}$ is implemented with three MLP layers, $E_{mask}^{i}$ is the mask embedding, $\operatorname{sigmoid}$ denotes a sigmoid operation, $\cdot$ denotes dot product operation, $\mathcal{M}_{clip}^i$ is the predicted masks of the entire clip. 

\begin{figure}[tb]
\begin{center}
\includegraphics[width=0.9\linewidth]{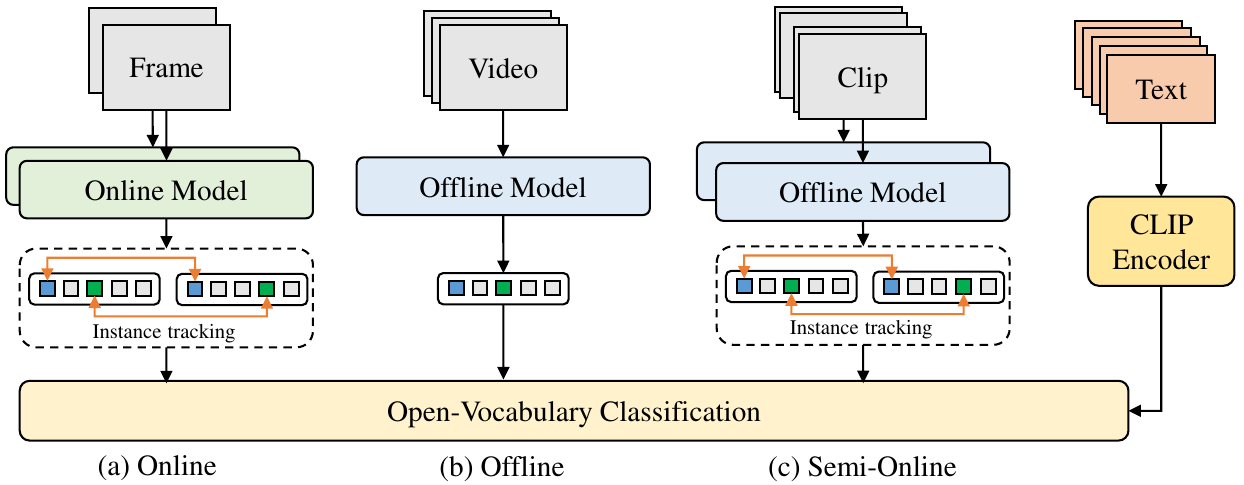}
\end{center}
\caption{Illustration of different inference schemes. (a) Online inference takes a single frame as input and associates instances frame by frame; (b) Offline inference takes the entire video as input, without associating instances; (c) Semi-online inference takes video clips as input and associates instances clip by clip.}
\label{fig:track}
\end{figure}

\noindent\textbf{Instance Association via Semi-online Tracking.} 
Considering our OVFormer is trained on the video level, we thus could handle the VIS in an \textit{offline} manner (\cref{fig:track} (b)) that feeds the whole video sequence into the model and tracks instances implicitly without any post-processing to associate instances. However, it is difficult to represent all instances of the entire video using instance queries, especially for long video and multiple instances cases~\cite{meinhardt2023novis}. 

Therefore, we employ a more flexible setting called \textit{semi-online} inference scheme (\cref{fig:track} (c)) to mitigate these issues. In concrete, semi-online scheme uses video clip as input and performs clip-level instance tracking with the Hungarian algorithm~\cite{kuhn1955hungarian} during testing:
\begin{equation}
  \label{equ:clip track}
  S^i_{track}= \operatorname{cos}(q_{clip}^i, q_{clip}^{i+1}), 
\end{equation}
where $S^i_{track}$ is the similarity score matrix between the $i$-th video-level instance queries and $i+\!1$-th instance queries, $\operatorname{cos}(\cdot)$ denotes the cosine similarity. We argue that this semi-online inference scheme not only better utilizes multi-frame information compared to \textit{online} method (\cref{fig:track} (a)), but also makes it possible for our models to conduct long video inference compared to \textit{offline} setting.

\subsection{Model Training}
To guide the instance query association learning and the segmentation, we take full advantage of the supervision signal and design the following loss function:
\begin{equation}
    \begin{split}
    \mathcal{L}_{match}(\hat{y}^i,y^i)\!=\!&
    \lambda_{ins}\mathcal{L}_{ins}(\hat{S}^i_{ins},S_{ins}^i)\!+\! 
    \lambda_{cls}\mathcal{L}_{cls}(\hat{S}^i_{cls},S_{cls}^i) \\ 
    & + \lambda_{mask}\mathcal{L}_{mask}(\hat{\mathcal{M}}^i_{clip},\mathcal{M}_{clip}^i),
    \label{equ:all-loss}
    \end{split}
\end{equation}
where the first two items indicate instance loss $\mathcal{L}_{ins}$ and classification loss $\mathcal{L}_{cls}$ that works for guiding the instance head (\cref{equ:instance}) and class head (\cref{equ:class}) learning, respectively.  We use the binary cross-entropy loss to implement these two loss functions.
The third item is mask loss $\mathcal{L}_{clip}$ (\cref{equ:mask}) which is the sum of dice loss~\cite{milletari2016v} and binary cross-entropy loss. The ground truth $y^i$ consists of an instance label $S^i_{ins}$, a class label $S^i_{cls}$ and segmentation mask $\mathcal{M}^i_{mask}$ for each instance.
We assign the ground truth set $y^i$ to the prediction set $\hat{y}^i$ by minimizing the total loss function $\mathcal{L}_{match}$.

\subsection{Implementation Details}
\label{sec:impl}
\noindent\textbf{Model architecture.} We implement our model using the Detectron2~\cite{wu2019detectron2} framework based on Mask2Former~\cite{cheng2022masked}. We experiment with both ResNet50~\cite{he2016deep} and SwinB~\cite{liu2021swin} as backbones. The text and image encoder of CLIP use the ViT-B/32 by default, and the parameters are frozen during training. 
The number of instance queries $N$ is set to 100. 

\noindent\textbf{Training and inference details.} We set $\lambda_{ins}$=2, $\lambda_{cls}$=2, $\lambda_{mask}$=5 as the weight of each loss (\cref{equ:all-loss}). We use AdamW~\cite{loshchilov2017decoupled} optimizer and the step learning rate schedule. We use an initial learning rate of 0.0001 and a weight decay of 0.05 for all backbones. A learning rate multiplier of 0.1 is applied to the backbone and we decay the learning rate at 0.9 and 0.95 fractions of the total number of training steps by a factor of 10. 
We first pre-train OVFormer on LVIS~\cite{gupta2019lvis} for 30 epochs with a batch size of 8 on 4 NVIDIA 3090 GPUs as OV2Seg~\cite{wang2023towards}. Then, we perform video-based training on LV-VIS~\cite{wang2023towards}. We decay the learning rate at 1/2 fraction of the total number of training steps by a factor of 10. We train OVFormer for 2k iterations with a batch size of 8 for LV-VIS.  For each video, we sample $T=2$ frames randomly
to build a video clip. 

During semi-online reference, we process each testing video in a
sequential manner by dividing the video into non-overlapping clips with fixed lengths. We resize the shorter side of each frame to 360 pixels for ResNet50~\cite{he2016deep} and to 480 pixels for SwinB~\cite{liu2021swin}. 

\section{Experiments}

\subsection{Datasets and Metrics}
We train OVFormer on the union of common and frequent categories in LVIS~\cite{gupta2019lvis} and base categories in LV-VIS~\cite{wang2023towards} and evaluate the performance on multiple video instance segmentation datasets, including LV-VIS~\cite{wang2023towards}, YouTube-VIS 2019~\cite{yang2019video}, YouTube-VIS 2021 and OVIS~\cite{qi2022occluded}.

\begin{itemize}
\item\noindent \textbf{LVIS} is a commonly employed open-vocabulary image detection dataset, comprising an extensive collection of 1,203 distinct categories. Following ViLD~\cite{gu2021open}, we take  866 frequent and common categories as the base~(training) categories and hold out the 337 rare categories as the novel categories.

\item\noindent \textbf{LV-VIS} is a large vocabulary video instance segmentation dataset, which comprises 4,828 real-world videos from 1,196 categories. All videos are divided into 3,083 training videos, 837 validation videos and 908 test videos. The categories are split into 641 base categories inherited from frequent and common categories in LVIS~\cite{gupta2019lvis}, and 555 novel categories disjoint with the base categories.

\item\noindent \textbf{YouTube-VIS 2019} contains 2,883 YouTube videos with 2,238 training videos, 302 validation videos and 343 test videos. It has a 40-category label set and 131k high-quality manual annotations. The categories are split into 33 base categories and 7 novel categories following the partitions in LVIS~\cite{gupta2019lvis}.

\item \noindent \textbf{YouTube-VIS 2021} is an improved and expanded version of the YouTube-VIS 2019 dataset, which includes more videos, improved categories, and doubled the number of annotations. In the same way, the categories are divided into 34 base categories and 6 novel categories.

\item \noindent \textbf{OVIS} is a very challenging dataset that contains long video sequences with a large number of objects and more frequent occlusion. It has 25 object categories, with only one category not in LVIS~\cite{gupta2019lvis} base categories; therefore, we only report the performance of the overall categories for OVIS.

\end{itemize}

\noindent \textbf{Evaluation Metrics.} We calculate the mean Average Precision (mAP) for all categories and then provide a more detailed breakdown, including mAP$_b$ for base categories and mAP$_n$ for novel categories.

\subsection{Main Results}
\label{sec:results}
In this section, we first test the Open-Vocabulary VIS ability of OVFormer on the recently released LV-VIS~\cite{wang2023towards} dataset. Then, we report the zero-shot generalization performance of OVFormer on traditional video instance segmentation datasets: YouTube-VIS 2019~\cite{yang2019video}, YouTube-VIS 2021, and OVIS~\cite{qi2022occluded}.
\begin{table}[t]
\setlength\tabcolsep{3.5pt}
\caption{The performance comparison on \textbf{LV-VIS} validation and test set. OV2Seg has not been fine-tuned on LV-VIS. For fair comparison, * means that we perform per-frame fine-tuning on LV-VIS using the weights provided by the authors.}
\centering
\scalebox{0.9}{
\begin{tabular}{l c c c c c c c c}
\toprule
\multirow{2}{*}{Method}&\multirow{2}{*}{Backbone}&\multirow{2}{*}{Training}&\multicolumn{3}{c}{Val
}&\multicolumn{3}{c}{Test}\\
\cmidrule(lr){4-6}
\cmidrule(lr){7-9}
&&&mAP&mAP$_b$&mAP$_n$&mAP&mAP$_b$&mAP$_n$\\
\midrule
DetPro~\cite{du2022learning}-SORT~\cite{bewley2016simple} &R50&LVIS&6.4&10.3&3.5&5.8&10.8&2.1\\
Detic~\cite{zhou2022detecting}-SORT~\cite{bewley2016simple} &R50&LVIS&6.5&10.7&3.4&5.7&10.6&2.1\\
DetPro~\cite{du2022learning}-OWTB~\cite{liu2022opening} &R50&LVIS&7.9&12.9&4.2&7.0&12.6&2.9\\
Detic~\cite{zhou2022detecting}-OWTB~\cite{liu2022opening} &R50&LVIS&7.7&12.6&4.2&7.0&12.8&2.8\\
Detic~\cite{zhou2022detecting}-XMem~\cite{cheng2022xmem} &R50&LVIS&8.8&13.4&5.4&7.7&13.3&3.6\\
OV2Seg~\cite{wang2023towards} &R50&LVIS&14.2&17.2&11.9&11.4&14.9&8.9\\
OV2Seg*&R50&LV-VIS&14.5&20.6&10.1&11.6&16.3&7.1\\
\textbf{OVFormer(Ours)}&R50&LV-VIS&\textbf{21.9}&\textbf{22.1}&\textbf{21.8}&\textbf{15.2}&\textbf{18.0}&\textbf{13.1}\\
\midrule
Detic~\cite{zhou2022detecting}-SORT~\cite{bewley2016simple} &SwinB&LVIS&12.8&21.1&6.6&9.4&15.8&4.7\\
Detic~\cite{zhou2022detecting}-OWTB~\cite{liu2022opening} &SwinB&LVIS&14.5&22.6&8.5&11.8&19.6&6.1\\
Detic~\cite{zhou2022detecting}-XMem~\cite{cheng2022xmem} &SwinB&LVIS&16.3&24.1&10.6&13.1&20.5&7.7\\
OV2Seg~\cite{wang2023towards} &SwinB&LVIS&21.1&\textbf{27.5}&16.3&16.4&\textbf{23.3}&11.5\\
\textbf{OVFormer(Ours)}&SwinB&LV-VIS&\textbf{24.7}&26.8&\textbf{23.1}&\textbf{19.5}&23.1&\textbf{16.7}\\
\bottomrule
\end{tabular}
}
\label{tab:LVVIS}
\end{table}

\begin{table}[t]
\caption{The performance comparison on \textbf{YouTube-VIS 2019}, \textbf{YouTube-VIS 2021} and \textbf{OVIS} validation sets. The \textbf{open} in the second column indicates whether the method is based on open-vocabulary.  For the \textbf{non-open} methods, we only report the overall mean average precision mAP as these methods can not handle novel categories. All the \textbf{open}-marked methods are trained on image dataset LVIS and evaluated on the video instance segmentation datasets directly, \ie, they are not fine-tuned using the training set of each dataset.}
\centering
\setlength\tabcolsep{2.5pt}
\scalebox{0.85}{
\begin{tabular}{l c c c c c c c c c c}
\toprule
\multirow{2}{*}{Method}&\multirow{2}{*}{Open}&\multirow{2}{*}{Backbone}&\multirow{2}{*}{Training}&\multicolumn{3}{c}{YTVIS2019}&\multicolumn{3}{c}{YTVIS2021}&\multicolumn{1}{c}{OVIS}\\
\cmidrule(lr){5-7}
\cmidrule(lr){8-10}
\cmidrule(lr){11-11}
&&&&mAP&mAP$_b$&mAP$_n$&mAP&mAP$_b$&mAP$_n$&mAP\\
\midrule
FEELVOS~\cite{voigtlaender2019feelvos} &\ding{55} & R50 & VIS & 26.9 &- &- &-&-&-&9.6 \\
MaskTrack~\cite{yang2019video} &\ding{55} & R50 & VIS & 30.3 &- &- &28.6&-&-&10.8\\
SipMask~\cite{cao2020sipmask} &\ding{55} &R50 & VIS & 33.7 & - & - & 31.7 & - & - &10.2\\
%VisTR &\ding{55}& R50 & 36.2 &  &  & -&&&&&&- \\
Mask2Former-VIS~\cite{cheng2021mask2former} &\ding{55}& R50 & VIS & 46.4 & - & - & 40.6 &-&-&17.3\\
%Mask2Former & SwinB & 59.5 & - & - & 52.0 &-&-&-&-&-&25.8\\
\midrule
\midrule
Detic~\cite{zhou2022detecting}-SORT~\cite{bewley2016simple}&\ding{51}&R50&LVIS&14.6&17.0&3.5&12.7&14.4&3.1&6.7\\
Detic~\cite{zhou2022detecting}-OWTB~\cite{liu2022opening}&\ding{51}&R50&LVIS&17.9&20.7&4.5&16.7&18.6&5.8&9.0\\
OV2Seg~\cite{wang2023towards}&\ding{51}&R50&LVIS&27.2&30.1&11.1&23.6&26.5&7.3&11.2\\
\textbf{OVFormer(Ours)}&\ding{51}&R50&LVIS&\textbf{34.8}&\textbf{38.7}&\textbf{16.5}&\textbf{29.8}&\textbf{32.3}&\textbf{15.7}&\textbf{15.1}\\
\midrule
Detic~\cite{zhou2022detecting}-SORT~\cite{bewley2016simple}&\ding{51}&SwinB&LVIS&23.8&27.2&7.9&21.6&23.7&9.8&11.7\\
Detic~\cite{zhou2022detecting}-OWTB~\cite{liu2022opening} &\ding{51}&SwinB&LVIS&30.0&34.3&9.7&27.1&29.9&11.4&13.6\\
OV2Seg~\cite{wang2023towards}&\ding{51}&SwinB&LVIS&37.6&41.1&21.3&33.9&36.7&18.2&17.5\\
\textbf{OVFormer(Ours)}&\ding{51}&SwinB&LVIS&\textbf{44.3}&\textbf{49.2}&\textbf{21.5}&\textbf{37.6}&\textbf{41.0}&\textbf{18.3}&\textbf{21.3}\\
\bottomrule
\end{tabular}
}
\label{tab:ytvis}
\end{table}

\noindent \textbf{Results on LV-VIS dataset.}
\cref{tab:LVVIS} reports the performance comparison between OVFormer and some baseline models on LV-VIS validation set, including OV2Seg~\cite{wang2023towards} and some propose-reduce-association baseline models~\cite{wang2023towards}. OVFormer with ResNet-50 backbone achieves 21.9 mAP, 22.1 mAP$_b$ and 21.8 mAP$_n$, outperforming the leading method OV2Seg by a large margin with \texttt{+}\textbf{7.7}, \texttt{+}\textbf{4.9} and \texttt{+}\textbf{9.9}, respectively. Compared to OV2Seg* under the same training setting with us, OVFormer still offers better performance, \ie, 22.1 \textit{vs.} 20.6 (mAP$_b$) for base categories and 21.8 \textit{vs.} 10.1 (mAP$_n$) for novel categories. Also, the performance of OVFormer with SwinB backbone further improves, reaching 24.7 mAP, 26.8 mAP$_b$ and 23.1 mAP$_n$.

Furthermore, we can observe that the segmentation results for novel categories are much inferior to the base categories for OV2Seg* (20.6 mAP$_b$ \textit{vs.} 10.1 mAP$_n$). In contrary (see \cref{fig:histogram}), our solution achieves approximate performance on base categories and novel categories (22.1 mAP$_b$ \textit{vs.} 21.8 mAP$_n$).  Especially for novel categories, our method boosts the performance of \textbf{83.2\%} compared to OV2Seg, which demonstrates the effectiveness of unified embedding alignment for novel categories identification. Therefore, our OVFormer yields strong generalization ability for novel categories compared to the existing baselines. 

\noindent \textbf{Zero-shot Generalization on VIS Datasets.}
To further evaluate the zero-shot generalization ability of our model, we directly test OVFormer on three close-vocabulary VIS datasets without fine-tuning the model using the training sample on the corresponding dataset.  Results in \cref{tab:ytvis} show that OVFormer achieves a stronger base-to-new generalization on both YouTube-VIS 2019 and YouTube-VIS 2021 benchmarks. In concrete, OVFormer with ResNet-50 backbone obtains 34.8 mAP, 38.7 mAP$_b$ and 16.5 mAP$_n$ on YouTube-VIS 2019, surpassing the main counterpart OV2Seg~\cite{wang2023towards} by \texttt{+}\textbf{7.6}, \texttt{+}\textbf{8.6} and \texttt{+}\textbf{5.4}. The same observation can also be seen on YouTubeVIS-2021, where OVFormer achieves 29.8 mAP, 32.3 mAP$_b$ and 15.7 mAP$_n$, showing significant overall improvement compared to OV2Seg. In addition, with SwinB backbone, OVFormer achieves 44.3 mAP on YouTube-VIS 2019 and 37.6 mAP on YouTube-VIS 2021 outperforming all baseline models.

Notably,  for the more challenging OVIS benchmark, all compared methods suffer from performance degradation due to the long-term videos and serious object occlusions. In spite of the presence of complex scenes, our method still delivers a solid overtaking trend: our OVFormer with ResNet-50 backbone still achieves the best score of 15.1 mAP, which is higher than OV2Seg \texttt{+}\textbf{3.9} mAP. Moreover, OVFormer with SwinB backbone sets a new state-of-the-art performance of 21.3 mAP, demonstrating the potential of our OVFormer in long and complicated video scenes. 

Considering the significant accuracy advantage of three datasets, the results demonstrate that OVFormer has a powerful zero-shot generalization ability. Notably, compared to the close-set fully-supervised methods, our method with ResNet 50 outperforms SipMask~\cite{cao2020sipmask} on YouTube-VIS 2019 (34.8 \textit{vs.} 33.7), MaskTrack R-CNN~\cite{yang2019video} on YouTube-VIS 2021 (29.8 \textit{vs.} 28.6), and even achieves comparable performance to Mask2Former-VIS~\cite{cheng2021mask2former} on OVIS (15.1 \textit{vs.} 17.3).

\subsection{Diagnostic Experiment}
For in-depth analysis, we conduct extensive ablation experiments on the LV-VIS and YouTube-VIS 2019 validation sets  to demonstrate the effectiveness of each component and the impact of different configurations. 

\noindent \textbf{Unified Embedding Alignment.} 
We first verify the effectiveness of the unified embedding alignment. We use OV2Seg~\cite{wang2023towards} as our baseline. As shown in \cref{tab:Abl}, unified embedding alignment provides an significant performance gain of \textbf{6.5} mAP on LV-VIS  and \textbf{7.6} mAP on YouTube-VIS 2019 compared to the baseline model. It proves that the proposed module is effective in conducting open-vocabulary classification by capturing better features. Especially for novel categories, the improvement brought by this module is more attractive (\texttt{+}\textbf{9.8} mAP$_n$ on LV-VIS). We attribute the performance gain to the interaction between the instance queries and CLIP image features, resulting in feature-aligned class embeddings, which greatly enhances the classification ability of open-vocabulary.

\begin{table}[t]
  \centering
  \setlength\tabcolsep{5.5pt}
  \caption{Ablation study of our contributions on LV-VIS and YTVIS2019 validation sets. We use OV2Seg as our \textit{Baseline}. Here, \textit{Baseline} and \textit{unified embedding alignment} variants adopt online inference, \textit{Video-based training} adopts offline inference. $\dag$ means the close-vocabulary VIS training setting, without deleting the novel category annotations in the training set.}
  \scalebox{0.85}{
  \begin{tabular}{lccccccc}
    \toprule
    \multirow{2}{*}{Variant}&\multirow{2}{*}{Training}&\multicolumn{3}{c}{LV-VIS}&\multicolumn{3}{c}{YTVIS2019}\\
    \cmidrule(lr){3-5}
    \cmidrule(lr){6-8}
    &&mAP&mAP$_b$&mAP$_n$&mAP&mAP$_b$&mAP$_n$\\
    \midrule
    Baseline & LVIS & 14.2 & 17.4 & 11.9 & 27.2 & 30.1 & 11.1\\
    +Unified embedding alignment & LVIS & 20.7 & 19.2 & 21.7 & 34.8 & 38.7 & 16.5\\
    \midrule
    +Video-based training & LV-VIS & 21.1 & 22.6 & 19.9 & - & - & -\\
    +Semi-online inference & LV-VIS & 21.9 & 22.1 & 21.8 & - & - & -\\
    \midrule
    +Video-based training & YTVIS19 & - & - & -& 44.1 & 49.7 & 18.0\\
    +Semi-online inference & YTVIS19 & - & - & -& 45.5 & 50.4 & 22.3\\
    \midrule
    \midrule
    Mask2Former-VIS~\cite{cheng2021mask2former} & YTVIS19\textsuperscript{$\dag$} & -& -& -& 46.4 & - & -\\
    OVFormer & YTVIS19\textsuperscript{$\dag$} & -& -& -& 51.4 & - & -\\
  \bottomrule
\end{tabular}
}
\label{tab:Abl}
\end{table}

\begin{figure}[th]
  \centering
    \begin{minipage}[b]{0.48\textwidth}
    \centering
    \includegraphics[width=\textwidth]{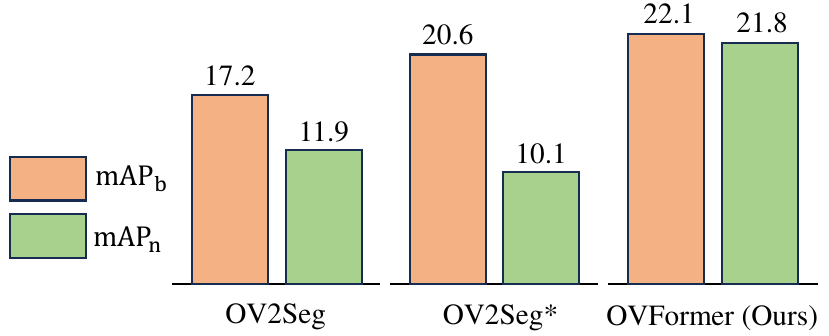}
    \caption{The performance comparison of base and novel categories on LV-VLIS validation set.}
    \label{fig:histogram}
  \end{minipage}
  \hfill
    \begin{minipage}[b]{0.48\textwidth}
    \centering
    \includegraphics[width=\textwidth]{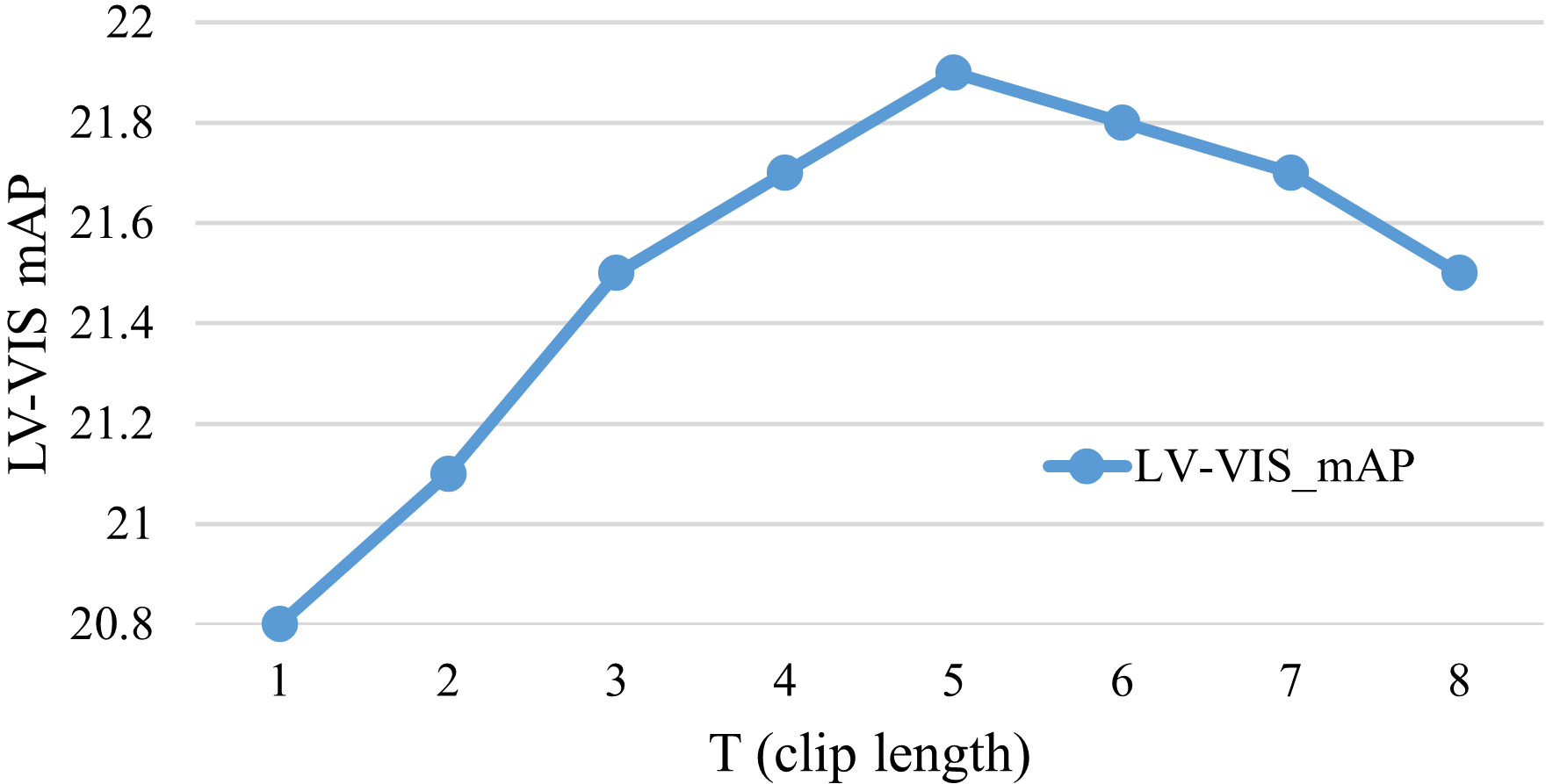}
    \caption{Ablation study of the clip length $T$ during OVFormer inference on LV-VIS validation set.}
    \label{fig:length}
  \end{minipage}
\end{figure}

\begin{table}[th]
  \centering
  \setlength\tabcolsep{10pt}
  \caption{Model parameters and FLOPs. Here, the backbone is ResNet50, Encoder is pixel decoder, Decoder is transformer decoder, UEA denotes the proposed unified embedding alignment module. The input image size is $800\times1333$.}
  \begin{tabular}{lcccc}
    \toprule
    Module & Backbone & Encoder & Decoder & UEA\\
    \midrule    
    Params & 23.5M & 6.0M & 14.5M & 1.2M\\
    FLOPs & 71.3G & 121.0G & 25.3G & 0.1G\\
  \bottomrule
\end{tabular}
\label{tab:param}
\end{table}

\noindent \textbf{Video-based Training.} 
As shown in \cref{tab:Abl}, we study the impact of video-based training scheme. Without modifying the architecture, the loss or even the training pipeline, we can see the performance improvement on YouTube-VIS 2019 in terms of mAP (\texttt{+}\textbf{9.3}), especially for base categories (\texttt{+}\textbf{11} mAP$_b$). This clear improvement is mainly attributed to the full utilization of the temporal information in a video clip, further enhancing temporal consistency of instances at minimal training cost. 

\noindent \textbf{Semi-online Inference.} 
By replacing the whole video with a clip-by-clip input type, the semi-online inference scheme further brings the mAP$_n$ from 19.9 to 21.8 on LV-VIS, which also boosts the performance gain of novel category on YouTube-VIS 2019 (18.0 $\rightarrow$ 22.3). In addition, we evaluate the performance of different inference clip lengths on LV-VIS in \cref{fig:length}. As the inference clip length increases, the optimal performance can be obtained when clip length $T$=5. 

\noindent \textbf{Close-set Training.} 
Finally, for more completed evaluation of our model, we test our model in the close-set setting. Without deleting the novel category labels in YouTube-VIS training set, our OVFormer achieves 51.4 mAP, significantly exceeding Mask2Former-VIS by \texttt{+}\textbf{5}. This indicates that the open-vocabulary model still has significant advantages in close-vocabulary tasks.

\noindent \textbf{Complexity analysis.}
We demonstrate parameters and FLOPs of OVFormer in \cref{tab:param}. We can see that UAE module reconciles the domain gap with very few parameters and FLOPs.

\begin{figure}[t]
\begin{center}
\includegraphics[width=1\linewidth]{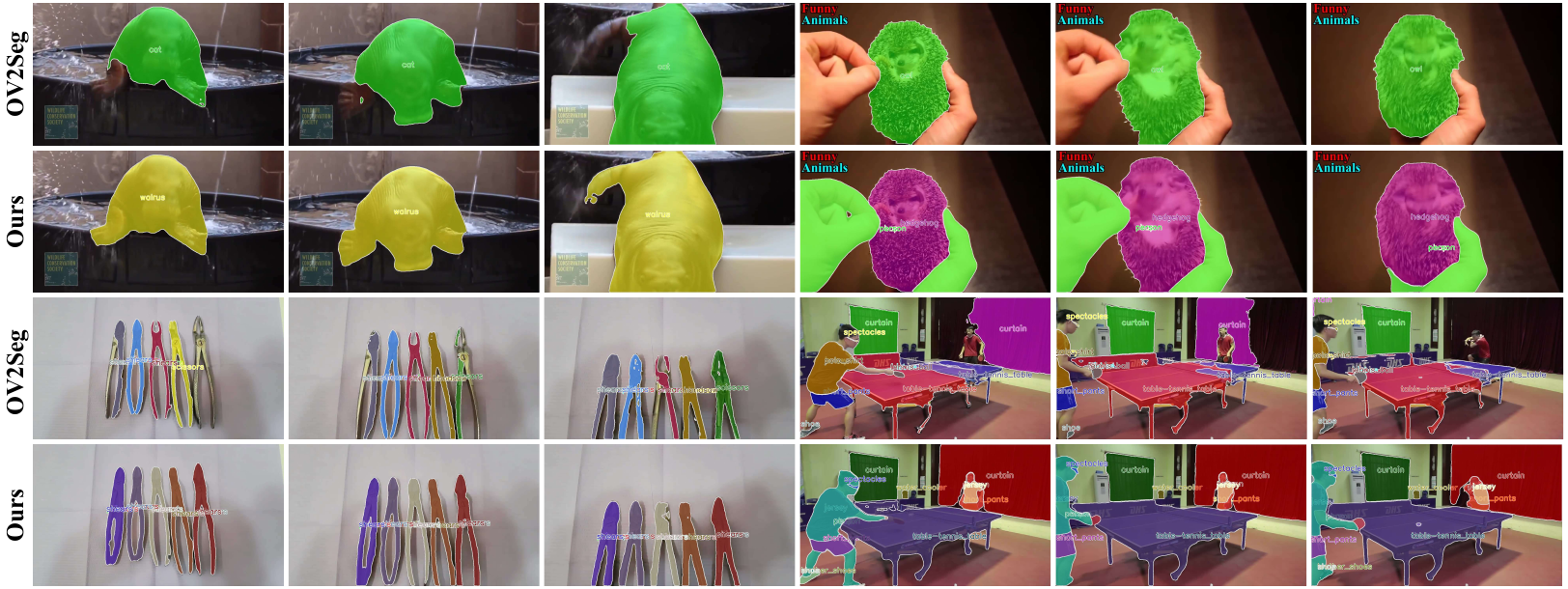}
\end{center}
\caption{Qualitative comparison of our OVFormer to baseline OV2Seg~\cite{wang2023towards} on the LV-VIS~\cite{wang2023towards} validation dataset. Both methods take the ResNet-50 as backbone. Each row shows two challenging video sequences.}
\label{fig:visual}
\end{figure}

\subsection{Qualitative Results} \cref{fig:visual} provides some representative qualitative results of ours against OV2Seg~\cite{wang2023towards} on the large-scale LV-VIS  set. 
The first row shows that OV2Seg is prone to misclassifying instances in novel categories into visually similar categories, such as \textit{walrus} is classified as \textit{cat} and \textit{hedgehog} is classified as \textit{owl}. OVFormer not only correctly classifies instances, but also provides clearer details for small parts in the second row. The third row shows the segmentation results of densely distributed instances with occlusion. OV2Seg fails to segment the instance across the video consistently, such as \textit{shears} and \textit{curtain}. However, OVFormer successfully segments and tracks all instances stably in these challenging scenes.

\section{Conclusion}
In this paper, we discover that the domain gap between the VLM features and the instance queries as well as the underutilization of temporal consistency are two central causes for the inadequate performance of the current Open-Vocabulary VIS method. We propose a novel end-to-end open-vocabulary video instance segmentation framework, named OVFormer. With the proposed unified embedding alignment, we obtain feature-aligned class embeddings, thus enhancing the generalization ability of novel categories.
We conduct video-level training without modifying the architecture and deploy semi-online inference scheme, which improves the temporal consistency of video instance segmentation. Extensive experimental results demonstrated that our method surpasses the state-of-the-art methods by a large margin. %We hope that our approach will serve as a strong baseline for future research in open-vocabulary video instance segmentation.

\section*{Acknowledgements}
This work was supported in part by the National Natural Science Foundation of China (No. 62106128, 62101309), the Natural Science Foundation of Shandong Province (No. ZR2021QF001, ZR2021QF109), Shandong Province Science and Technology Small and
Medium-sized Enterprise Innovation Capacity Enhancement Project (2023TSGC0115), Shandong Province Higher Education Institutions Youth Entrepreneurship and Technology Support Program (2023KJ027).

\bibliographystyle{splncs04}
\bibliography{main}

\begin{thebibliography}{10}
\providecommand{\url}[1]{\texttt{#1}}
\providecommand{\urlprefix}{URL }
\providecommand{\doi}[1]{https://doi.org/#1}

\bibitem{bewley2016simple}
Bewley, A., Ge, Z., Ott, L., Ramos, F., Upcroft, B.: Simple online and realtime tracking. In: ICIP. pp. 3464--3468. IEEE (2016)

\bibitem{cao2020sipmask}
Cao, J., Anwer, R.M., Cholakkal, H., Khan, F.S., Pang, Y., Shao, L.: Sipmask: Spatial information preservation for fast image and video instance segmentation. In: ECCV. pp. 1--18. Springer (2020)

\bibitem{carion2020end}
Carion, N., Massa, F., Synnaeve, G., Usunier, N., Kirillov, A., Zagoruyko, S.: End-to-end object detection with transformers. In: ECCV. pp. 213--229. Springer (2020)

\bibitem{chen2023open}
Chen, X., Li, S., Lim, S.N., Torralba, A., Zhao, H.: Open-vocabulary panoptic segmentation with embedding modulation. In: ICCV. pp. 1141--1150 (2023)

\bibitem{cheng2021mask2former}
Cheng, B., Choudhuri, A., Misra, I., Kirillov, A., Girdhar, R., Schwing, A.G.: Mask2former for video instance segmentation. arXiv preprint arXiv:2112.10764  (2021)

\bibitem{cheng2022masked}
Cheng, B., Misra, I., Schwing, A.G., Kirillov, A., Girdhar, R.: Masked-attention mask transformer for universal image segmentation. In: CVPR. pp. 1290--1299 (2022)

\bibitem{cheng2022xmem}
Cheng, H.K., Schwing, A.G.: Xmem: Long-term video object segmentation with an atkinson-shiffrin memory model. In: ECCV. pp. 640--658. Springer (2022)

\bibitem{ding2022decoupling}
Ding, J., Xue, N., Xia, G.S., Dai, D.: Decoupling zero-shot semantic segmentation. In: CVPR. pp. 11583--11592 (2022)

\bibitem{du2022learning}
Du, Y., Wei, F., Zhang, Z., Shi, M., Gao, Y., Li, G.: Learning to prompt for open-vocabulary object detection with vision-language model. In: CVPR. pp. 14084--14093 (2022)

\bibitem{fang2024learning}
Fang, H., Zhang, T., Zhou, X., Zhang, X.: Learning better video query with sam for video instance segmentation. TCSVT pp. 1--12 (2024)

\bibitem{gu2021open}
Gu, X., Lin, T.Y., Kuo, W., Cui, Y.: Open-vocabulary object detection via vision and language knowledge distillation. In: ICLR. pp. 1--14 (2021)

\bibitem{guo2023openvis}
Guo, P., Huang, T., He, P., Liu, X., Xiao, T., Chen, Z., Zhang, W.: Openvis: Open-vocabulary video instance segmentation. arXiv preprint arXiv:2305.16835  (2023)

\bibitem{gupta2019lvis}
Gupta, A., Dollar, P., Girshick, R.: Lvis: A dataset for large vocabulary instance segmentation. In: CVPR. pp. 5356--5364 (2019)

\bibitem{he2016deep}
He, K., Zhang, X., Ren, S., Sun, J.: Deep residual learning for image recognition. In: CVPR. pp. 770--778 (2016)

\bibitem{heo2022vita}
Heo, M., Hwang, S., Oh, S.W., Lee, J.Y., Kim, S.J.: Vita: Video instance segmentation via object token association. NeurIPS  \textbf{35},  23109--23120 (2022)

\bibitem{huang2022minvis}
Huang, D.A., Yu, Z., Anandkumar, A.: Minvis: A minimal video instance segmentation framework without video-based training. NeurIPS  \textbf{35},  31265--31277 (2022)

\bibitem{hwang2021video}
Hwang, S., Heo, M., Oh, S.W., Kim, S.J.: Video instance segmentation using inter-frame communication transformers. NeurIPS  \textbf{34},  13352--13363 (2021)

\bibitem{jia2021scaling}
Jia, C., Yang, Y., Xia, Y., Chen, Y.T., Parekh, Z., Pham, H., Le, Q., Sung, Y.H., Li, Z., Duerig, T.: Scaling up visual and vision-language representation learning with noisy text supervision. In: ICML. pp. 4904--4916. PMLR (2021)

\bibitem{kuhn1955hungarian}
Kuhn, H.W.: The hungarian method for the assignment problem. Naval research logistics quarterly  \textbf{2}(1-2),  83--97 (1955)

\bibitem{li2023tcovis}
Li, J., Yu, B., Rao, Y., Zhou, J., Lu, J.: Tcovis: Temporally consistent online video instance segmentation. In: ICCV. pp. 1097--1107 (2023)

\bibitem{li2023ovtrack}
Li, S., Fischer, T., Ke, L., Ding, H., Danelljan, M., Yu, F.: Ovtrack: Open-vocabulary multiple object tracking. In: CVPR. pp. 5567--5577 (2023)

\bibitem{lin2021video}
Lin, H., Wu, R., Liu, S., Lu, J., Jia, J.: Video instance segmentation with a propose-reduce paradigm. In: ICCV. pp. 1739--1748 (2021)

\bibitem{liu2022opening}
Liu, Y., Zulfikar, I.E., Luiten, J., Dave, A., Ramanan, D., Leibe, B., O{\v{s}}ep, A., Leal-Taix{\'e}, L.: Opening up open world tracking. In: CVPR. pp. 19045--19055 (2022)

\bibitem{liu2021swin}
Liu, Z., Lin, Y., Cao, Y., Hu, H., Wei, Y., Zhang, Z., Lin, S., Guo, B.: Swin transformer: Hierarchical vision transformer using shifted windows. In: ICCV. pp. 10012--10022 (2021)

\bibitem{loshchilov2017decoupled}
Loshchilov, I., Hutter, F.: Decoupled weight decay regularization. arXiv preprint arXiv:1711.05101  (2017)

\bibitem{lu2020deep}
Lu, X., Ma, C., Shen, J., Yang, X., Reid, I., Yang, M.H.: Deep object tracking with shrinkage loss. PAMI  \textbf{44}(5),  2386--2401 (2020)

\bibitem{lu2020zero}
Lu, X., Wang, W., Shen, J., Crandall, D., Luo, J.: Zero-shot video object segmentation with co-attention siamese networks. PAMI  \textbf{44}(4),  2228--2242 (2020)

\bibitem{lu2021segmenting}
Lu, X., Wang, W., Shen, J., Crandall, D.J., Van~Gool, L.: Segmenting objects from relational visual data. PAMI  \textbf{44}(11),  7885--7897 (2021)

\bibitem{ma2022open}
Ma, Z., Luo, G., Gao, J., Li, L., Chen, Y., Wang, S., Zhang, C., Hu, W.: Open-vocabulary one-stage detection with hierarchical visual-language knowledge distillation. In: CVPR. pp. 14074--14083 (2022)

\bibitem{meinhardt2023novis}
Meinhardt, T., Feiszli, M., Fan, Y., Leal-Taixe, L., Ranjan, R.: Novis: A case for end-to-end near-online video instance segmentation. arXiv preprint arXiv:2308.15266  (2023)

\bibitem{milletari2016v}
Milletari, F., Navab, N., Ahmadi, S.A.: V-net: Fully convolutional neural networks for volumetric medical image segmentation. In: 3DV. pp. 565--571. IEEE (2016)

\bibitem{qi2022occluded}
Qi, J., Gao, Y., Hu, Y., Wang, X., Liu, X., Bai, X., Belongie, S., Yuille, A., Torr, P.H., Bai, S.: Occluded video instance segmentation: A benchmark. IJCV  \textbf{130}(8),  2022--2039 (2022)

\bibitem{radford2021learning}
Radford, A., Kim, J.W., Hallacy, C., Ramesh, A., Goh, G., Agarwal, S., Sastry, G., Askell, A., Mishkin, P., Clark, J., et~al.: Learning transferable visual models from natural language supervision. In: ICML. pp. 8748--8763. PMLR (2021)

\bibitem{vaswani2017attention}
Vaswani, A., Shazeer, N., Parmar, N., Uszkoreit, J., Jones, L., Gomez, A.N., Kaiser, {\L}., Polosukhin, I.: Attention is all you need. NeurIPS  \textbf{30},  1--11 (2017)

\bibitem{voigtlaender2019feelvos}
Voigtlaender, P., Chai, Y., Schroff, F., Adam, H., Leibe, B., Chen, L.C.: Feelvos: Fast end-to-end embedding learning for video object segmentation. In: CVPR. pp. 9481--9490 (2019)

\bibitem{wang2023towards}
Wang, H., Yan, C., Wang, S., Jiang, X., Tang, X., Hu, Y., Xie, W., Gavves, E.: Towards open-vocabulary video instance segmentation. In: ICCV. pp. 4057--4066 (2023)

\bibitem{wang2021actionclip}
Wang, M., Xing, J., Liu, Y.: Actionclip: A new paradigm for video action recognition. arXiv preprint arXiv:2109.08472  (2021)

\bibitem{wang2021end}
Wang, Y., Xu, Z., Wang, X., Shen, C., Cheng, B., Shen, H., Xia, H.: End-to-end video instance segmentation with transformers. In: CVPR. pp. 8741--8750 (2021)

\bibitem{wu2023betrayed}
Wu, J., Li, X., Ding, H., Li, X., Cheng, G., Tong, Y., Loy, C.C.: Betrayed by captions: Joint caption grounding and generation for open vocabulary instance segmentation. In: ICCV. pp. 21938--21948 (2023)

\bibitem{wu2022seqformer}
Wu, J., Jiang, Y., Bai, S., Zhang, W., Bai, X.: Seqformer: Sequential transformer for video instance segmentation. In: ECCV. pp. 553--569. Springer (2022)

\bibitem{wu2022defense}
Wu, J., Liu, Q., Jiang, Y., Bai, S., Yuille, A., Bai, X.: In defense of online models for video instance segmentation. In: ECCV. pp. 588--605. Springer (2022)

\bibitem{wu2019detectron2}
Wu, Y., Kirillov, A., Massa, F., Lo, W.Y., Girshick, R.: Detectron2. \url{https://github.com/facebookresearch/detectron2} (2019)

\bibitem{yang2019video}
Yang, L., Fan, Y., Xu, N.: Video instance segmentation. In: ICCV. pp. 5188--5197 (2019)

\bibitem{yang2022temporally}
Yang, S., Wang, X., Li, Y., Fang, Y., Fang, J., Liu, W., Zhao, X., Shan, Y.: Temporally efficient vision transformer for video instance segmentation. In: CVPR. pp. 2885--2895 (2022)

\bibitem{yang2024scalable}
Yang, Z., Miao, J., Wei, Y., Wang, W., Wang, X., Yang, Y.: Scalable video object segmentation with identification mechanism. PAMI pp. 1--15 (2024)

\bibitem{ying2023ctvis}
Ying, K., Zhong, Q., Mao, W., Wang, Z., Chen, H., Wu, L.Y., Liu, Y., Fan, C., Zhuge, Y., Shen, C.: Ctvis: Consistent training for online video instance segmentation. In: ICCV. pp. 899--908 (2023)

\bibitem{zareian2021open}
Zareian, A., Rosa, K.D., Hu, D.H., Chang, S.F.: Open-vocabulary object detection using captions. In: CVPR. pp. 14393--14402 (2021)

\bibitem{zhang2023dvis}
Zhang, T., Tian, X., Wu, Y., Ji, S., Wang, X., Zhang, Y., Wan, P.: Dvis: Decoupled video instance segmentation framework. In: CVPR. pp. 1282--1291 (2023)

\bibitem{zhou2022extract}
Zhou, C., Loy, C.C., Dai, B.: Extract free dense labels from clip. In: ECCV. pp. 696--712. Springer (2022)

\bibitem{zhou2022detecting}
Zhou, X., Girdhar, R., Joulin, A., Kr{\"a}henb{\"u}hl, P., Misra, I.: Detecting twenty-thousand classes using image-level supervision. In: ECCV. pp. 350--368. Springer (2022)

\end{thebibliography}

\end{document}